%% file: main.tex
\documentclass[11pt]{article}

\usepackage[final]{acl}

\usepackage{times}
\usepackage{latexsym}

\usepackage[T1]{fontenc}

\usepackage[utf8]{inputenc}

\usepackage{microtype}

\usepackage{inconsolata}

\usepackage{graphicx}

\usepackage{xcolor}
\usepackage{amssymb}
\usepackage{amsxtra}
\usepackage{multirow}
\usepackage{diagbox}
\usepackage{longtable}
\usepackage{tabularx}
\usepackage[table]{xcolor}
\usepackage{array}
\usepackage{makecell}
\usepackage{stfloats}
\usepackage{balance}
\usepackage{multicol}
\usepackage{amsmath, bm}
\usepackage{booktabs} 
\usepackage{arydshln}
\usepackage{xspace}
\usepackage{enumitem}
\usepackage{threeparttable}
\usepackage{dsfont}
\usepackage{hyperref}
\usepackage{bbm}
\usepackage{makecell}
\usepackage[linesnumbered,ruled,vlined]{algorithm2e}
\usepackage{algorithmic}
\usepackage{color} 
\usepackage{subfigure}
\usepackage{stfloats}
\usepackage{subcaption}
\usepackage{mdframed}

\usepackage{listings}
\usepackage{appendix}
\usepackage{titletoc}
\usepackage{amsthm}
\usepackage{float}
\usepackage{cuted}
\usepackage[most]{tcolorbox} 

\newcommand{\method}{Sparse-RL\xspace}
\newcommand{\vpara}[1]{\vspace{0.05in}\noindent \textbf{#1 }}

\title{Sparse-RL: Breaking the Memory Wall in LLM Reinforcement Learning via Stable Sparse Rollouts}

\author{
\normalfont
Sijia Luo$^{1,2}$\thanks{Equal Contributions.},
Xiaokang Zhang$^{1,2}$\footnotemark[1],
Yuxuan Hu$^{1,2}$,
Bohan Zhang$^{1,2}$,\\
Ke Wang$^{1,2}$,
Jinbo Su$^{1,2}$,
Mengshu Sun$^{3}$,
Lei Liang$^{3}$,
Jing Zhang$^{1,2}$\thanks{Corresponding Author.}
\\[3pt]
$^1$School of Information, Renmin University of China, Beijing, China \\
$^2$Key Laboratory of Data Engineering and Knowledge Engineering, Beijing, China \\
$^3$Ant Group, Hangzhou, China
\\[1pt]
\small \texttt{sijialuo23@gmail.com, \{zhang2718, zhang-jing\}@ruc.edu.cn}
}

\begin{document}
\maketitle

\input{latex/0.abstract}

\section{Introduction}
\label{sec:introduction}
\input{latex/1.introduction}

\section{Related Work}
\label{sec:related}
\input{latex/2.related}

\section{Problem Definition}
\label{sec:problem}
\input{latex/3.problem}

\section{Methodology}
\label{sec:method}
\input{latex/4.method}

\section{Experiments}
\label{sec:experiment}
\input{latex/5.exp}

\section{Conclusion}
\label{sec:conclusion}
\input{latex/6.conclusion}

\section*{Limitations}
\label{sec:limitations}
\input{latex/7.limitations}


\bibliography{custom}

\clearpage

\appendix
\label{sec:appendix}
\input{latex/8.appendix}

\end{document}

%% file: latex/0.abstract.tex
\begin{abstract}
Reinforcement Learning (RL) has become essential for eliciting complex reasoning capabilities in Large Language Models (LLMs). However, the substantial memory overhead of storing Key-Value (KV) caches during long-horizon rollouts acts as a critical bottleneck, often prohibiting efficient training on limited hardware. While existing KV compression techniques offer a remedy for inference, directly applying them to RL training induces a severe policy mismatch, leading to catastrophic performance collapse. To address this, we introduce \method, which empowers stable RL training under sparse rollouts. We show that instability arises from a fundamental policy mismatch among the dense old policy, the sparse sampler policy, and the learner policy. To mitigate this issue, \method incorporates Sparsity-Aware Rejection Sampling and Importance-based Reweighting to correct the off-policy bias introduced by compression-induced information loss. Experimental results show that \method reduces rollout overhead compared to dense baselines while preserving the performance. Furthermore, \method inherently implements sparsity-aware training, significantly enhancing model robustness during sparse inference deployment. The corresponding training data and code are
publicly available on the repository\footnote{\url{https://github.com/RUCKBReasoning/Sparse-RL}}.
 
\end{abstract}

%% file: latex/1.introduction.tex
Reasoning large models, such as OpenAI-o1~\cite{jaech2024openai} and DeepSeek-R1~\cite{guo2025deepseek}, have demonstrated exceptional capabilities in tackling complex reasoning tasks such as mathematical reasoning and code generation~\cite{chen2025towards, liu2025code}. Reinforcement learning (RL) has emerged as an effective post-training paradigm for enhancing LLMs' strong reasoning capabilities~\cite{ouyang2022training, shao2024deepseekmath}. By incentivizing models to explore extended chains of reasoning and optimizing them toward explicit behavioral goals, RL enables LLMs to better align with human preferences and execute tasks requiring structured reasoning and verifiable decision-making~\cite{guo2025deepseek, team2025qwq}.

The remarkable effectiveness of RL training comes at an extremely high computational cost. Modern RL training primarily consists of two stages: rollout and training, with the rollout stage dominating the overall time in RL execution. Recent studies~\cite{shao2025beat, wang2025rlhfspec} show that the rollout stage accounts for approximately 70\% of the total time. 
Simultaneously, this large-scale generation faces a severe ``memory wall'': as the sequence lengths increase, the continuously expanding KV cache consumes substantial GPU memory~\cite{zhao2025accelerating}. To prevent out-of-memory (OOM) errors during long-tail sample generation, rollout batch sizes must be constrained. This limitation reduces GPU utilization and constrains the throughput of scalable RL training.

\begin{figure}[t]
  \centering
  \includegraphics[width=0.49\textwidth]{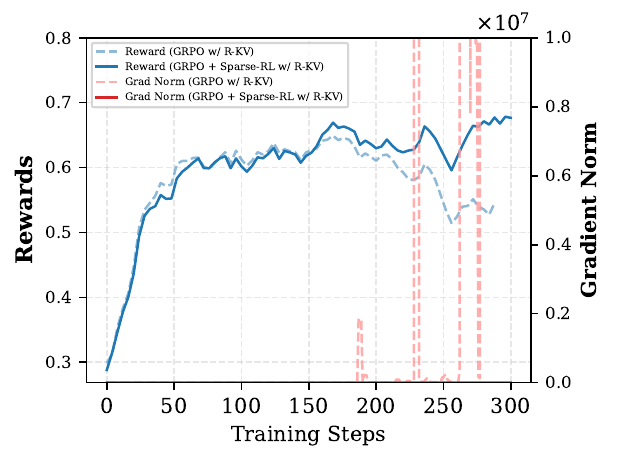}
  \caption{Impact of \method on training stability. We compare training dynamics between naive GRPO and \method during Qwen2.5-7B training with R-KV compression. While naive GRPO suffers from a reward collapse and gradient spikes, \method maintains stable reward growth with gradient norms remaining consistently low (appearing near the x-axis relative to the massive spikes of the naive baseline).}
 \label{fig:collapse}
\end{figure}

To alleviate computational and memory bottlenecks in RL training, KV cache compression techniques offer a viable solution—originally designed to accelerate long-context inference~\cite{zhang2023h2o}. By selectively retaining key-value pairs (typically identified through cumulative attention scores), representative compression algorithms (e.g., H2O~\cite{zhang2023h2o}, StreamingLLM~\cite{xiao2023efficient}, SnapKV~\cite{li2024snapkv}) can effectively reduce memory complexity from linear growth to a fixed token budget, decoupling memory consumption from sequence length. Given that existing training-free compression methods demonstrate strong performance in inference scenarios, a natural question emerges: \textit{Can these training-free compression methods be directly applied to the rollout stage of RL training?} Theoretically, this could eliminate the aforementioned long-tail memory shortage risk, thereby supporting larger batch sizes to accelerate rollout.

However, directly applying compression methods designed for static inference to dynamic RL training is non-trivial and faces the following challenges: (1) \textit{Policy Mismatch}: In standard PPO~\cite{schulman2017proximal} or GRPO~\cite{shao2024deepseekmath}, sampling and training policies typically originate from identical distributions. In this setup, responses generated during sampling rely on a sparse policy based on compressed key-value pairs, while gradient updates utilize a dense policy. This mismatch introduces a fundamental off-policy issue~\cite{yao2025offpolicy, liu-li-2025}.  (2) \textit{Anomalous Sample}: Existing KV compression algorithms are primarily designed for long-context inference and do not account for the randomness inherent in the RL exploration stage. During sampling, the information loss induced by compression may generate anomalous samples such as infinite repetition (example in Appendix \ref{sec:example}), which can produce massive negative gradients or noise, undermining the stability of RL training. Crucially, these two effects jointly result in catastrophic training collapse. The policy mismatch amplifies the influence of corrupted trajectories, while anomalous samples introduce extreme gradients and high-variance updates. As observed in Figure \ref{fig:collapse}, after training for about 200 steps, the rewards of the naive GRPO with R-KV~\cite{cai2025r} compression begin to decrease and the gradient norm spikes at the same time.

To address these challenges, we propose \method, a novel framework that enables stable RL training with sparse rollouts through KV cache compression. For anomalous samples generated due to KV compression, we design Sparsity-Aware Rejection Sampling, which utilizes the token probability ratio between sparse and dense policies as a detector to eliminate anomalous sequence samples. 
To further correct the policy mismatch within valid samples, we use Importance-based Reweighting which leverage the token probability ratio between sparse and dense policies to reweight the policy gradient, recovering unbiased estimates for the dense learner.
This design enables \method to theoretically achieve stable, efficient RL training under relatively low memory budgets. Since its correction mechanism relies solely on probability distribution rather than specific compression operators, \method adapts to various mainstream training-free compression algorithms, providing the community with a universal, memory-efficient RL paradigm.


Our contributions can be summarized as follows: 1) We pioneer the exploration of integrating training-free KV Cache compression technology into the rollout phase of RL training. 2) We identified and resolved training collapse caused by policy mismatch resulting from sparse rollouts. We propose Sparsity-Aware Rejection Sampling to filter anomalous trajectories introduced by KV compression, and Importance-based Reweighting to ensure stable policy updates. 3) We evaluate \method on 7 benchmarks with 4 models and 2 KV compression methods. Under a fixed token budget, \method matches dense rollouts and even outperforms them on Qwen2.5-1.5B. Up to 7B scale, it retains over 96\% of dense performance. 4) We further observe that \method inherently implements \textit{Sparsity-Aware Training}, significantly enhancing model robustness during sparse inference deployment in downstream tasks. This successfully bridges the gap between efficient training and efficient deployment.

%% file: latex/2.related.tex

\vpara{Efficient Reinforcement Learning for LLM.}
Reinforcement Learning (RL) has emerged as the cornerstone for eliciting complex reasoning capabilities in LLMs, as demonstrated by OpenAI-o1~\cite{jaech2024openai} and DeepSeek-R1~\cite{guo2025deepseek}. 
Unlike offline alignment methods such as DPO~\cite{rafailov2023direct}, which are data-efficient but lack exploratory capacity, online algorithms like PPO~\cite{schulman2017proximal} and GRPO~\cite{shao2024deepseekmath} are essential for discovering novel reasoning paths. Compared to PPO, GRPO eliminates the critic by estimating advantages through group-wise ranking, significantly reducing memory consumption while maintaining strong performance on verifiable reasoning tasks. 
GSPO~\cite{zheng2025group} elevates the optimization granularity from token-level to sequence-level, rectifying the credit assignment ambiguity inherent in dense rewards. From a system perspective, frameworks like Seer~\cite{qin2025seeronlinecontextlearning}, AgentRL~\cite{zhang2025agentrl} and AReaL~\cite{fu2025areal} propose asynchronous architectures that decouple trajectory generation from policy learning, significantly improve training throughput.

\vpara{KV Cache Compression.} Existing methods primarily include token eviction~\cite{ge2023model, liu2023scissorhands, li2024snapkv}, merging~\cite{zhang2024cam}, quantization~\cite{hooper2024kvquant, liu2024kivi}, and low-rank decomposition~\cite{sun2024shadowkv}. For reasoning models, we consider dynamic eviction strategies that continuously evicts key-value caches during decoding. Early methods like StreamingLLM~\cite{xiao2023efficient} use sliding windows combined with attention sinks. SnapKV~\cite{li2024snapkv}, PyramidKV~\cite{cai2024pyramidkv} and Ada-KV~\cite{feng2024ada} prune tokens based on attention scores. More recently, R-KV~\cite{cai2025r} targets redundant tokens common in reasoning chains. By eliminating redundancy via similarity clustering during both pre-filling and decoding, R-KV achieves state-of-the-art results on MATH500~\cite{hendrycks2021measuring} and AIME24. However, all aforementioned methods are designed for inference, and their application to RL training dynamics has not yet been investigated.

%% file: latex/3.problem.tex
In the context of RL training with KV cache compression, we aim to optimize a policy $\pi_\theta$ to maximize the expected objective. Due to sparse rollout generation and policy staleness, three distinct policy distributions coexist during training, giving rise to an intrinsic policy mismatch.

\vpara{\textbf{Notation}} We define an autoregressive LLM parameterized by $\theta$ as a policy $\pi_{\theta}$. We use $x$ to denote an input prompt and $D$ as the prompt set. Under the policy $\pi_{\theta}$, the likelihood of a response $o$ to a prompt $x$ is denoted
as $\pi_{\theta}(o|x) = \prod_{t=1}^{|o|} \pi_{\theta}(o_t|x, o_{<t})$ where $|o|$ is the number of tokens in $o$.

\vpara{\textbf{Dense Old Policy}} represents the ideal and uncompressed policy derived from the full context history $o_{<t}$:
\begin{equation}
    \pi_{\theta_\text{old}}(o_t | x,o_{<t}) = P(o_t | x, o_{<t}; \theta_{\text{old}})
\end{equation}

\vpara{\textbf{Sparse Sampler Policy}} is the policy actually used for generating responses, emerging from the sparse version of the dense old policy. Due to the use of the KV compression operator $\mathcal{M(\cdot)}$, the conditional probability at time step $t$ depends on compressed historical information:
\begin{equation}
    \pi_{\theta_\text{sparse}}(o_t | x, o_{<t}) = P(o_t | x,\mathcal{M}( o_{<t}); \theta_{\text{old}})
\end{equation}
Note that while we denote it as $\pi_{\theta_\text{sparse}}$ for notational symmetry, it does not possess a distinct set of parameters; rather, it represents the sparse view of $\pi_{\theta_\text{old}}$ under compression.
  
\vpara{\textbf{Learner Policy}} is the policy currently being optimized, typically diverges from $\theta_{\text{old}}$ due to policy staleness:
\begin{equation}
    \pi_{\theta}(o_t | x, o_{<t}) = P(o_t | x, o_{<t}; \theta)
\end{equation}

%% file: latex/4.method.tex
To address the policy mismatch introduced by KV compression, we first analyze the discrepancy from an Importance Sampling (IS) perspective. 
Guided by this derivation, we propose \method, which utilizes Sparsity-Aware Rejection Sampling to filter anomalous trajectories and Importance-based Reweighting to recover unbiased gradient estimates.

\subsection{The Decomposition of Policy Mismatch}
Standard on-policy RL algorithms (e.g., PPO) assume the sampler policy matches the learner policy. In our setting, sampling from $\pi_{\theta_\text{sparse}}$ introduces a structural bias. To obtain an unbiased gradient estimate for $\pi_\theta$, one would theoretically employ the total importance sampling (IS) weight $w_{\text{total}} = \frac{\pi_{\theta}}{\pi_{\theta_\text{sparse}}}$. 

We observe that the discrepancy between the learner and the sparse sampler policy stems from two distinct sources: the difference between dense and sparse policy distributions caused by information loss during compression, and the standard off-policy bias arising from optimizing new policy based on empirical data generated by old policy. Therefore, the importance weights can be decomposed as follows:
\begin{equation}
\label{eq:decomposition}
\resizebox{0.95\linewidth}{!}{$\displaystyle
\frac{\pi_{\theta}(o_t | x, o_{<t})}{\pi_{\theta_{\text{sparse}}}(o_t | x, o_{<t})} = 
\underbrace{\frac{\pi_{\theta}(o_t | x, o_{<t})}{\pi_{\theta_{\text{old}}}(o_t | x, o_{<t})}}_{\text{policy staleness}} \times 
\underbrace{\frac{\pi_{\theta_{\text{old}}}(o_t | x, o_{<t})}{\pi_{\theta_{\text{sparse}}}(o_t | x, o_{<t})}}_{\text{sparsity-induced mismatch}}
$}
\end{equation}

\subsection{Sparsity-Aware Rejection Sampling}
\label{subsec:cars}

\vpara{Quantifying Distributional Discrepancy.}
KV compression can be viewed as a lossy approximation operator $\mathcal{M}(\cdot)$ on the historical state $o_{<t}$. While often accurate, this approximation causes $\pi_{\theta_\text{sparse}}$ to occasionally assign high probability to tokens that are statistically impossible under the full-context $\pi_{\theta_\text{old}}$, leading to abnormal sequences.
To detect these trajectories, we define the sparsity consistency ratio $\xi_t$ at each time step $t$:
\begin{equation}
\xi_t = \frac{\pi_{\theta_\text{old}}(o_t | x, o_{<t})}{\pi_{\theta_\text{sparse}}(o_t | x, o_{<t})}
\end{equation}
A value of $\xi_t \approx 1$ implies local consistency between sparse policy and dense policy. Conversely, $\xi_t \to 0$ signals a support mismatch, where the sparse policy enters a region of the state space unsupported by the dense model.

\vpara{Rejection Sampling.}
In reasoning tasks, a single hallucinated step can invalidate the entire chain of thought. Therefore, we use a strict sequence-level validity constraint: a trajectory is accepted if and only if every generated token remains within the support of the dense policy. We define the sequence-level rejection weight $M^{\text{RS}}(o)$ as:
\begin{equation}
    M^{\text{RS}}(o) = 
    \begin{cases} 
    0 & \text{if } \exists t \in [1, |o|] \text{ s.t. } \xi_t < \epsilon \\
    1 & \text{otherwise}
    \end{cases}
\end{equation}
where $\epsilon$ is a threshold (e.g., 1e-4).
This implies that if the sparse policy generates a single anomalous token, the weight of the entire sequence is zeroed out. As a result, the optimizer computes gradients only from logically valid trajectories that remain consistent with the behavior of the dense policy.

\subsection{Importance-based Reweighting}
\label{subsec:optimization}
As \method is built upon GRPO (preliminaries detailed in Appendix \ref{sec:preliminaries}), the policy model samples a group of $G$ outputs $\{o_1, \dots, o_G\}$ from the sparse policy $\pi_{\theta_{\text{sparse}}}$ for a given prompt $x$. After filtering out hallucinated trajectories through rejection sampling, a distributional bias remains: the valid samples are drawn from $\pi_{\theta_\text{sparse}}$, but we aim to estimate expectations under the dense policygg $\pi_{\theta_\text{old}}$.
To rigorously correct this bias, we incorporate the sparsity consistency ratio $\xi_{i,t} = \frac{\pi_{\theta_\text{old}}(o_t | x, o_{i,<t})}{\pi_{\theta_\text{sparse}}(o_t | x, o_{i,<t})}$ as a reweighting factor, and $w_{i,t}(\theta) = \frac{\pi_{\theta}(o_t | x, o_{i,<t})}{\pi_{\theta_\text{old}}(o_t | x, o_{i,<t})}$ denote the ratio of policy staleness.


The objective function of \method is formulated as:
\begin{equation}
\label{eq:final_objective_grpo}
\resizebox{0.99\linewidth}{!}{
$
\begin{aligned}
\mathcal{J}_{\text{Sparse-RL}}(\theta) = 
& \mathbb{E}_{\substack{x \sim \mathcal{D} \\ o_i \sim \pi_{\theta_{\text{sparse}}}}} 
\Bigg[ \frac{1}{G} \sum_{i=1}^{G} M^{\text{RS}}(o_i) \cdot \frac{1}{|o_i|} \sum_{t=1}^{|o_i|} \xi_{i,t} \\ 
& \cdot \min \Big( w_{i,t}(\theta) \hat{A}_{i},  \text{clip}\big(w_{i,t}(\theta), 1-\epsilon, 1+\epsilon\big) \hat{A}_{i} \Big)  \Bigg]
\end{aligned}
$}
\end{equation}
where $M^{\text{RS}}(o)$ zeros out entire anomalous trajectories. The sparsity consistency ratio $\xi_{i,t}$ is applied outside the clipping operator to unbiasedly correct the distribution shift via importance sampling, while the clipping is restricted to $w_{i,t}(\theta)$ to enforce the trust region constraint relative to the dense policy.

\vpara{Gradient Analysis}
For the \method objective (Eq. \ref{eq:final_objective_grpo}), the policy gradient $\nabla_\theta \mathcal{J}_{\text{Sparse-RL}}(\theta)$ can be derived as follows (clipping is omitted for brevity):
\begin{equation}
\resizebox{0.95\linewidth}{!}{
$
\begin{aligned}
\nabla_\theta \mathcal{J}_{\text{Sparse-RL}} = \operatorname*{\mathbb{E}}_{\substack{x \sim \mathcal{D} \\ o_i \sim \pi_{\theta_{\text{sparse}}}}} 
& \Bigg[ \frac{1}{G} \sum_{i=1}^{G} M^{\text{RS}}(o_i) \cdot \frac{\hat{A}_{i}}{|o_i|} \\ 
& \cdot \sum_{t=1}^{|o_i|} \Big( \xi_{i,t} \cdot w_{i,t}(\theta) \nabla_\theta \log \pi_\theta(o_{i,t}|x,o_{i,<t}) \Big) \Bigg]
\end{aligned}
$}
\end{equation}

If we use $\mathcal{V} = \{i \mid M^{\text{RS}}(o_i) = 1\}$ to denote the set of indices for valid trajectories that passed the rejection sampling filter, the gradient becomes:
\begin{equation}
\resizebox{0.95\linewidth}{!}{
$
\begin{aligned}
\nabla_\theta \mathcal{J}_{\text{Sparse-RL}} = \operatorname*{\mathbb{E}}_{\substack{x \sim \mathcal{D} \\ o_i \sim \pi_{\theta_{\text{sparse}}}}} 
&\Bigg[ \frac{1}{G} \sum_{i \in \mathcal{V}} \frac{1}{|o_i|} \cdot \hat{A}_{i} \\
& \cdot \sum_{t=1}^{|o_i|} \Big(\xi_{i,t} \cdot w_{i,t}(\theta)\nabla_\theta \log \pi_\theta(o_{i,t}|x,o_{i,<t}) \Big) \Bigg]
\end{aligned}
$}
\end{equation}
The detailed derivation is provided in Appendix \ref{sec:derivation}.

\vpara{Mechanistic Interpretation.}
The resulting gradient reveals a dual-level correction mechanism:
\begin{itemize}
\item \textbf{Sequence-level Filtering ($M^{\text{RS}}$):} The summation $\sum_{i \in \mathcal{V}}$ effectively removes any output $o_i$ that contains hallucinations or abnormal reasoning steps caused by information loss. This ensures the advantage estimates $\hat{A}_{i}$ are not corrupted by out-of-distribution samples.
\item \textbf{Token-level Reweighting ($\xi_{i,t}$):} For the accepted trajectories, the token-level gradient is reweighted by $\xi_{i,t}$ to eliminate the distribution gap between sparse and dense policy. 
\end{itemize}

%% file: latex/5.exp.tex
\subsection{Experimental Setup}
\input{table/main_result}

\vpara{Backbone Models} We conduct zero RL training experiments on Qwen2.5 (1.5B/3B/7B)~\cite{qwen2025qwen25technicalreport} and Llama-3.2-1B-Instruct~\cite{dubey2024llama}, to cover various model sizes and verify cross-architecture generalization.

\vpara{Training Datasets} 
Our experiments use the SimpleRL-Zoo~\cite{zeng2025simplerl} dataset, which integrates the training sets of GSM8K~\cite{cobbe2021training} and MATH~\cite{hendrycks2021measuring}. This dataset is chosen based on the observation that, when initiating ZeroRL directly from a base model, successful training critically depends on using data that matches the model’s capability. The dataset is divided into three difficulty levels: Easy (GSM8K and MATH level 1), Medium (MATH levels 1–4), and Hard (MATH levels 3–5), with each split containing approximately 8,000 problems. Given the strong mathematical capabilities of the Qwen2.5 series and llama3.2 instruct models, we use the hard level split as our training dataset.

\vpara{Evaluation} We evaluate the performance of our method on 7 standard mathematical reasoning benchmarks: GSM8K~\cite{cobbe2021training}, MATH500~\cite{hendrycks2021measuring}, Gaokao~\cite{zhang2023evaluating}, Minerva Math~\cite{lewkowycz2022solving}, OlympiadBench~\cite{he2024olympiadbench}, AIME24\footnote{huggingface.co/datasets/AI-MO/aimo-validation-aime} and AMC23\footnote{huggingface.co/datasets/AI-MO/aimo-validation-amc}. For AIME24 and AMC23, we report Avg@32: for each problem, we sample 32 responses and calculate the mean accuracy among these 32 responses; the final score is the average of the mean accuracy for each item over the dataset. For the other six benchmarks, we report Pass@1: generating one response per problem.

\vpara{Baselines} To assess the efficacy of our proposed framework, we design a comparative study involving three primary configurations:
\begin{itemize}
    \item \textbf{Base:} The base model without RL training.
    \item \textbf{GRPO (Dense Rollout):} The standard GRPO algorithm using full KV cache rollouts without any compression. This serves as the dense baseline to measure the performance gap caused by sparse approximation.
    \item \textbf{GRPO (Naive Sparse Rollout):} We directly apply the KV cache compression methods (specifically R-KV and SnapKV) during the rollout generation phase of GRPO, devoid of off-policy correction mechanisms. This baseline highlights the catastrophic performance collapse arising from the policy mismatch.
    \item \textbf{GRPO + \method (Ours):} We instantiate our \method framework with two distinct compression algorithms: \textbf{\method + SnapKV} (representing attention-based compression) and \textbf{\method + R-KV} (representing redundancy-aware compression).
\end{itemize}

\begin{figure*}[t]
  \centering
  \includegraphics[width=\textwidth]{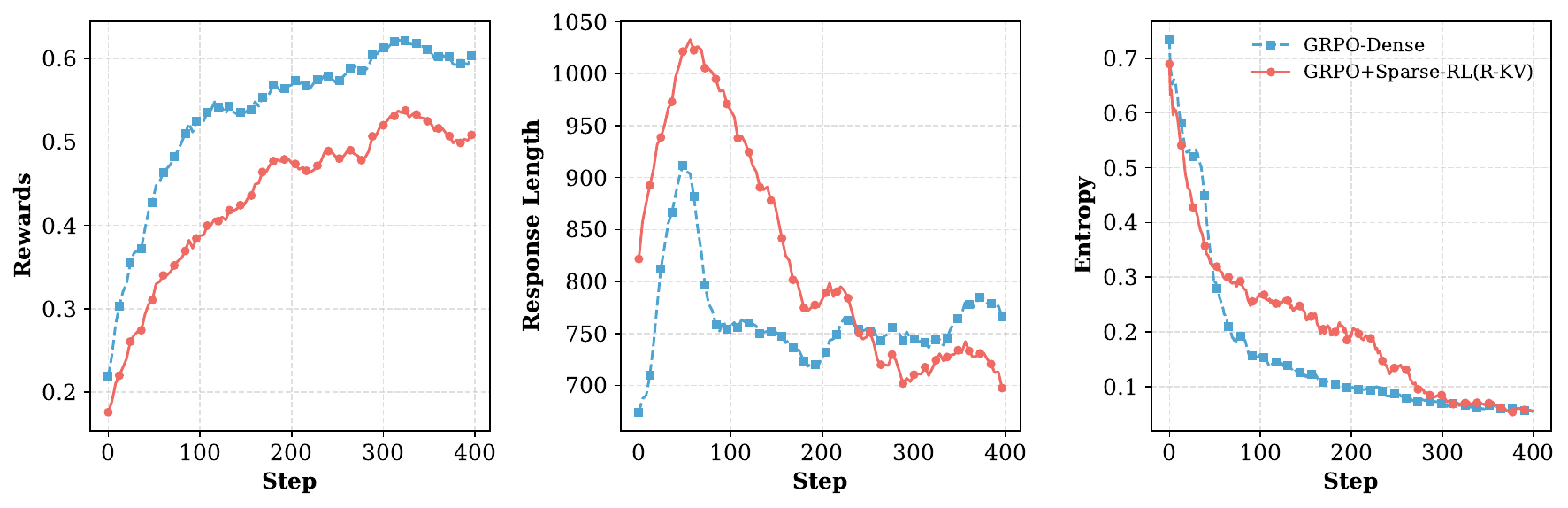}
  \caption{Comparison of training curves between GRPO-Dense and GRPO + \method (with R-KV) on Qwen2.5-3B over 400 training steps. \textbf{Left}: Average rewards per step. \textbf{Middle}: Average response length (tokens). \textbf{Right}: Policy entropy.}
 \label{fig:dynamics}
\end{figure*}

\vpara{Implementation Details} Our training is based on the slime framework~\cite{slime_github}. We adopt a strict binary reward scheme, assigning a reward of 1 to correct responses and 0 otherwise. For each prompt, we sample $G = 8$ rollouts with a temperature of 1.0, top-p of 1.0, and a maximum response length of 4096 tokens. The global rollout batch size is set to 1024, with an update batch size of 256. The learning rate is set to 1e-6 with a KL loss coefficient of 1e-4. The ratio of rejection sampling is set to 1e-4. As for compression parameters, we set the KV budget to 512, which means during the rollout process, only 512 tokens are retained when generating each response. Experiments fine-tune backbone models for 400 training steps ($\approx 1.5$ epochs) on $4\times$ NVIDIA H20-HGX-141GB GPUs.

\subsection{Main Results}
Table \ref{tab:main_result} presents the experimental results across 7 mathematical reasoning benchmarks. The ``Toks. saving'' column indicates the reduction in stored KV cache tokens compared to the generation length of the GRPO-Dense during RL training. Our key findings are as follows:

\vpara{\method achieves comparable performance to dense upper bounds.} 
Under the constraint of a fixed retained token budget, GRPO with naive sparse rollout performs poorly overall, while GRPO + \method maintains performance comparable to the Dense GRPO baseline and even outperforms it in certain scenarios. On Qwen2.5-1.5B, \method with R-KV achieves an average score of 36.2, outperforming GRPO-Dense (35.4) by 2.3\%, and attains the best results on 4 out of 7 benchmarks (GSM8K, Minerva, Olympiad, AMC23). On Qwen2.5-3B, \method performs on par with the dense baseline (42.8 vs. 43.4 on average). On Qwen2.5-7B, although a small performance gap remains (51.4 vs. 53.1 on average), \method retains 96.8\% of the dense model’s performance, indicating that the method scales effectively to larger models with more complex reasoning patterns.

\vpara{Substantial reduction in memory overhead.} 
As indicated by the ``Toks. saving'' metric, we achieve token savings of 35.1\%, 53.3\%, 42.0\%, and 39.4\% for the Llama3.2-1B-Instruct, Qwen2.5-1.5B, 3B, and 7B models, respectively. Notably, on the 1.5B model, GRPO + \method outperformed the dense GRPO method while discarding over half of the contextual history (53.3\% saving). This highlights the potential redundancy in standard CoT paths, showing that \method enables the model to perform correct deductions using a much more compact information representation.

\vpara{Robustness across models and compression algorithms.}
Consistent performance trends on both Qwen2.5 and Llama backbones indicate that our framework is robust across model architectures. Comparing two compression variants, \method with R-KV and \method with SnapKV, we observe highly similar results. For example, on Qwen2.5-3B, their average scores are nearly identical (42.8 vs. 42.6 on average). This indicates that \method is largely compression-agnostic: as long as the compression method preserves the essential context, our off-policy correction mechanism can effectively stabilize training. Notably, R-KV shows a slight advantage on more challenging tasks such as Olympiad (27.7 vs. 25.9 on 3B, 36.6 vs. 32.9 on 7B), validating its design benefits in handling reasoning redundancy.

\subsection{Training Dynamics}

To further analyze the training stability and potantial of \method, we visualize the training dynamics of GRPO-Dense and GRPO + \method (with R-KV) in Figure \ref{fig:dynamics}. Our analysis is as follows:

\vpara{Outcome Rewards. }
The average reward trajectory of \method is slightly lower than that of the dense baseline but remains within an acceptable margin, which is a direct consequence of the information loss inherent to compressed rollout generation. Despite the approximation errors introduced by KV compression, the reward curves indicate that \method can still perform stable and effective policy optimization. While a persistent yet minor performance gap remains compared to the GRPO-Dense baseline—which represents a theoretical upper bound—\method successfully avoids the catastrophic training collapse commonly observed in naive sparse training settings.

\begin{figure}[t]
  \centering
  \includegraphics[width=1.0\columnwidth]{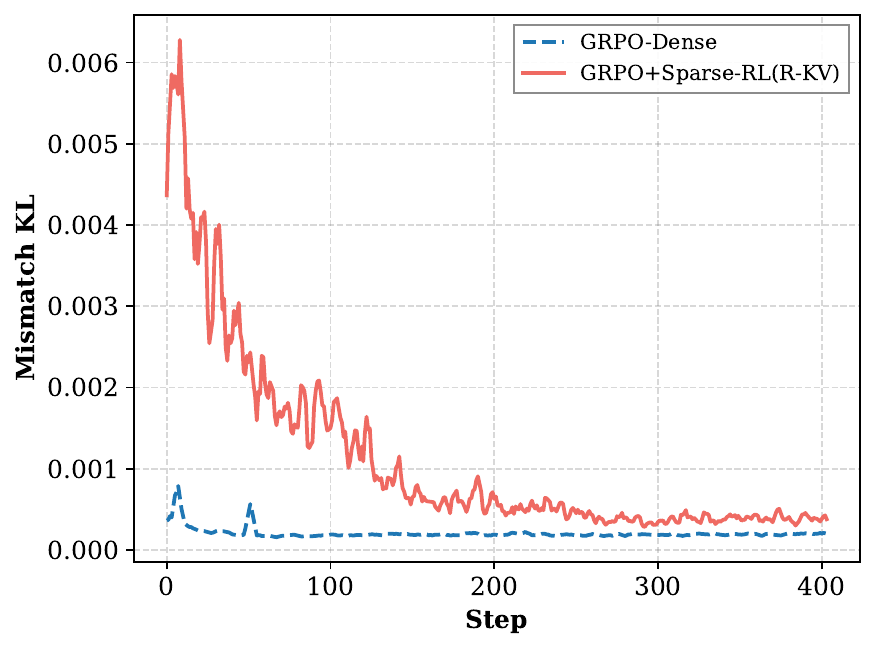}
  \caption{Comparison of the Mismatch KL between GRPO-Dense and GRPO + \method (with R-KV) on Qwen2.5-7B.}
 \label{fig:mis_k3_kl}
\end{figure}

\vpara{Adaptation to Sparse Memory.}
The overall trend in response length for \method (R-KV) and GRPO-Dense is to increase initially and then decrease. In the early stages of training, \method (R-KV) has longer response length, possibly because the model generated long anomalous sequences (e.g., infinite repetitions) due to information loss. Crucially, as training progresses, the response length rapidly regresses to align with the Dense baseline. This convergence validates the corrective power of our RL framework, demonstrating that the model learns to perform token-efficient reasoning within the strict KV budget constraints.

\vpara{Stability of Exploration.}
The GRPO-Dense baseline exhibits a steep drop in entropy during the initial phase, while \method demonstrates a more gradual decay curve. Notably, \method consistently maintains higher entropy levels between steps 50 and 300. We attribute this phenomenon to the implicit regularization effect of compression noise, which prevents the policy from prematurely overconfidently committing to shortcuts, thereby promoting exploration of the inference space. The entropy of \method ultimately converges to the same level as the dense baseline, confirming that our method successfully achieves policy stability and avoids the divergence issues commonly encountered in sparse training.

\vpara{Degree of Policy Mismatch.}
We quantify policy mismatch via the KL divergence between rollout and training policies. As shown in Figure \ref{fig:mis_k3_kl}, Sparse-RL initially exhibits a higher KL magnitude $10^{-3}$ than the dense baseline $10^{-4}$, reflecting the structural bias from KV compression. This discrepancy steadily decreases and stabilizes after approximately 200 steps. This convergence demonstrates that the dense learner successfully adapts to the compression logic, effectively bridging the distribution gap to learn consistent reasoning patterns from sparse trajectories.

\vpara{Sampling Efficiency and Stability.} 
To provide a more comprehensive view of our rejection sampling mechanism, we monitor the rejection ratio (the fraction of trajectories vetoed by the sparsity-aware filter) and policy gradient clipping ratio on Qwen2.5-3B throughout GRPO + \method (R-KV) training. The average rejection rate is 0.07, indicating that the majority of generated trajectories satisfy the sparsity consistency constraint. The average clip ratio remains at 0.0005. The detailed dynamics are provided in Appendix \ref{sec:ratio}.

\subsection{Bonus: Superiority in Sparse Inference}
\input{table/bonus}

Beyond the standard evaluation, we investigate the models' adaptability to downstream sparse inference scenarios. In this experiment, we enforce the exact same KV compression configuration used during \method training (i.e., R-KV with a budget of 512 tokens) for models trained by the GRPO-Dense and our \method (with R-KV). As shown in Table \ref{tab:bonus}, \method outperform their dense counterparts on the majority of benchmarks across varying scales. For instance, on Qwen2.5-3B, \method surpasses the dense baseline by 7.6\% on MATH500 under the sparse setting. The overall superiority highlights a critical insight: while GRPO-Dense suffers from a distribution shift when forcing sparsity at inference time, \method successfully internalizes the compression logic during training. This \textit{Sparsity-Aware Training} effectively aligns the training dynamics with the inference constraints, yielding significantly robust performance in memory-constrained applications.

\subsection{Ablation Study}
\begin{figure}[t]
  \centering
  \includegraphics[width=1.0\columnwidth]{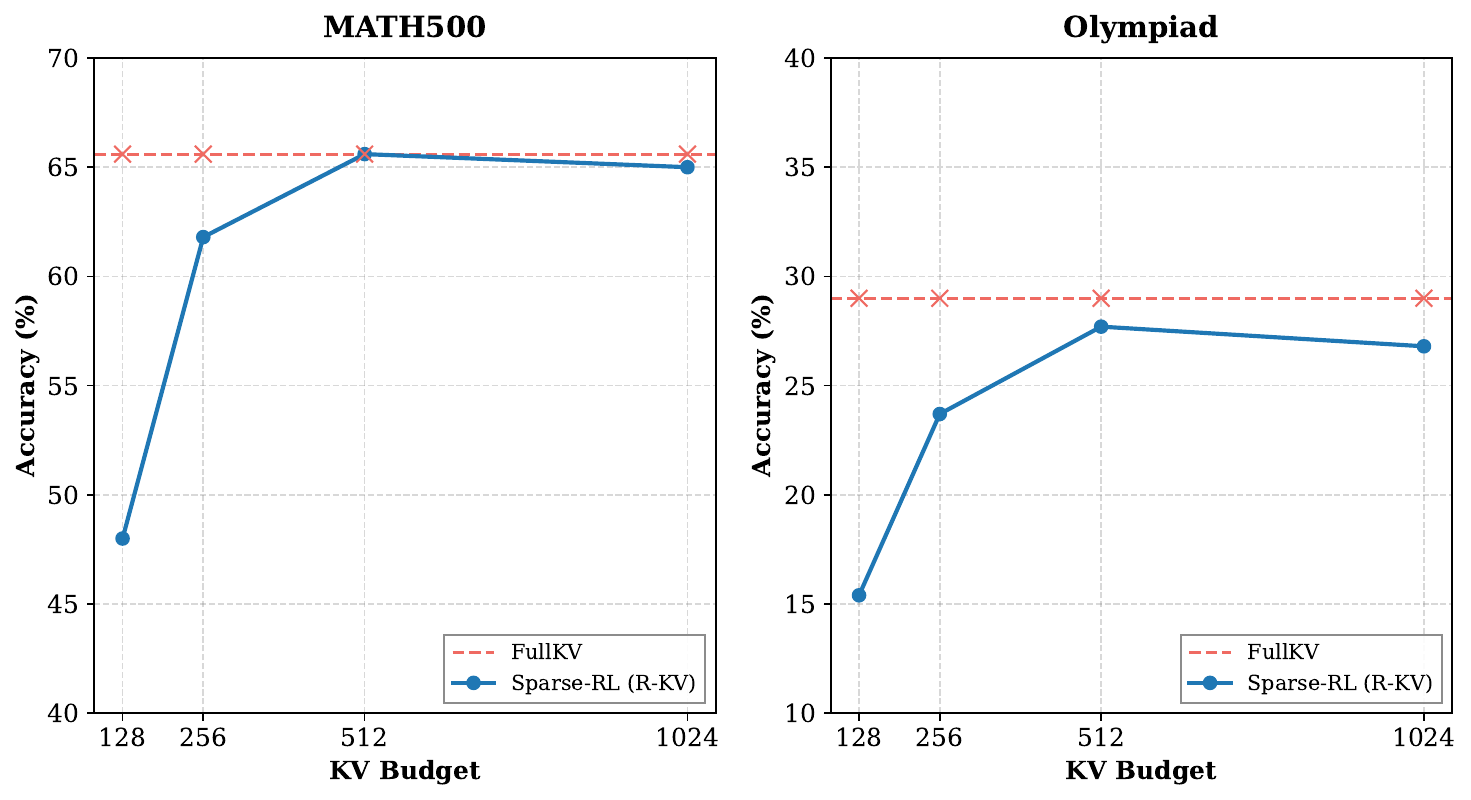}
  \caption{Ablation study on KV cache budget size. Performance of \method (with R-KV) on Qwen2.5-3B across different budget levels on MATH500 and Olympiad-Bench. The red dashed line represents the dense rollouts baseline (FullKV).}
 \label{fig:budget}
\end{figure}

\vpara{Size of KV Cache Budget.} We evaluate the effect of different token budget sizes by varying the retained tokens during rollout (128, 256, 512, 1024, and FullKV) for Qwen2.5-3B trained with GRPO + \method (using R-KV).
As shown in Figure \ref{fig:budget}, performance degrades under extremely small budgets (128 tokens), indicating insufficient context for reliable reasoning. As the budget increases to 256 and 512 tokens, accuracy improves rapidly on both benchmarks. We note that for the models and benchmarks considered in this paper, a KV budget of 512 tokens is sufficient to achieve performance comparable to FullKV. For more challenging tasks or models with longer CoT generation, the optimal budget may need to be adjusted accordingly.

%% file: table/main_result.tex
\begin{table*}[!ht] 
  \centering
  \footnotesize
  \begin{tabular}{@{}
    >{\raggedright}p{0.13\textwidth}   
    >{\centering}p{0.09\textwidth}   
    >{\centering\arraybackslash}p{0.056\textwidth}  
    >{\centering\arraybackslash}p{0.07\textwidth}   
    >{\centering\arraybackslash}p{0.05\textwidth}   
    >{\centering\arraybackslash}p{0.06\textwidth}   
    >{\centering\arraybackslash}p{0.068\textwidth}  
    >{\centering\arraybackslash}p{0.056\textwidth}  
    >{\centering\arraybackslash}p{0.056\textwidth}  
    >{\centering\arraybackslash}p{0.045\textwidth}   
    >{\centering\arraybackslash}p{0.055\textwidth}  
    @{}}
    \toprule

    \textbf{Model} & \textbf{Rollout Method} & \textbf{GSM8K} & \textbf{MATH500} & \textbf{Gaokao} & \textbf{Minerva Math} &
    \textbf{Olympiad} & \textbf{AIME24} & \textbf{AMC23} & \textbf{Avg.} & \textbf{Toks. saving} \\

    \midrule
    \rowcolor{gray!20}
    \multicolumn{11}{c}{\textbf{Llama-3.2-1B-Instruct}} \\
    \midrule
    
    Base & -- & 36.2 & 22.8 & 17.9 & 5.5 & 5.5 & 1.2 & 8.8 & 14.0 & - \\
    GRPO & Dense & 51.2 & \underline{33.6} & \textbf{28.8} & \textbf{8.8} & 8.7 & \underline{2.9} & 12.8 & \underline{21.0} & - \\
    
    \cmidrule(lr){1-11}
    GRPO & \multirow{2}{*}{w/ R-KV}
      & 49.8 & 30.0 & 23.9
      & 5.5 & 7.6 & 1.7
      & 12.8 & 18.8
      & \multirow{4}{*}{35.1\%} \\

    $\hookrightarrow$ + Sparse-RL & 
      & 49.1 & 30.2 & 26.5 & 5.5 & 8.3 
      & 2.1 & \textbf{15.2} & 19.6 & \\
    
    \cmidrule(lr){1-10}
    
    GRPO & \multirow{2}{*}{w/ SnapKV}
      & \underline{52.2} & \underline{33.6} & 26.5
      & 5.5 & \underline{9.0} & \underline{2.9}
      & \underline{13.8} & 20.5
      &  \\

    $\hookrightarrow$ + Sparse-RL & 
      & \textbf{54.6} & \textbf{34.4} & \underline{27.0} & \underline{6.2} & \textbf{9.3}
      & \textbf{3.0} & 13.6 & \textbf{21.2} & \\

    \midrule
    \rowcolor{gray!20}
    \multicolumn{11}{c}{\textbf{Qwen2.5-1.5B}} \\
    \midrule
    
    Base & -- & 43.5 & 21.0 & 18.4 & 4.0 & 4.0 & 0.3 & 7.5 & 14.1 & - \\
    GRPO & Dense & 72.9 & \textbf{58.0} & 46.5 & 16.9 & \underline{21.6} & \textbf{3.6} & 28.5 & 35.4 & - \\
    
    \cmidrule(lr){1-11}
    
    GRPO & \multirow{2}{*}{w/ R-KV}
    & 48.8 & 34.0 & 32.5 & 7.0 & 11.6 & 1.0 & 21.2 & 22.3 & \multirow{4}{*}{53.3\%} \\
    
    $\hookrightarrow$ + Sparse-RL & 
    & \textbf{75.3} & 55.8 & \underline{48.1} & \textbf{20.2} & \textbf{21.9} & 2.9 & \underline{29.5} & \textbf{36.2} & \\
    
    \cmidrule(lr){1-10}
    
    GRPO & \multirow{2}{*}{w/ SnapKV}
    & 66.3 & 37.6 & 33.8 & 8.1 & 16.1 & 3.1 & 18.0 & 26.1 & \\

    $\hookrightarrow$ + Sparse-RL & 
    & \underline{73.7} & \underline{57.6} & \textbf{48.8} & \underline{18.0} & 20.7 & \underline{3.4} & \underline{29.5} & \underline{36.0} & \\
    
    \midrule
    \rowcolor{gray!20}
    \multicolumn{11}{c}{\textbf{Qwen2.5-3B}} \\
    \midrule
    
    Base & -- & 76.0 & 55.8 & 45.2 & 19.9 & 21.6 & 4.1 & 27.3 & 35.7 & - \\
    GRPO & Dense & \underline{83.5} & \underline{64.2} & \textbf{56.9} & 25.0 & \textbf{29.8} & \textbf{6.0} & \underline{38.6} & \textbf{43.4} & - \\
    
    \cmidrule(lr){1-11}
    
    GRPO & \multirow{2}{*}{w/ R-KV}
    & 79.0 & 56.6 & 46.2 & 21.7 & 23.0 & 4.7 & 27.0 & 36.9 & \multirow{4}{*}{42.0\%} \\

    $\hookrightarrow$ + Sparse-RL & 
    & \textbf{84.6} & \textbf{65.6} & 53.5 & \textbf{26.5} & \underline{27.7} & 4.6 & 37.4 & \underline{42.8} & \\
    
    \cmidrule(lr){1-10}
    
    GRPO & \multirow{2}{*}{w/ SnapKV}
    & 79.0 & 54.2 & 48.3 & 19.5 & 23.6 & 5.2 & 30.5 & 37.2 &  \\
    
    $\hookrightarrow$ + Sparse-RL & 
    & 83.4 & 64.0 & \underline{54.8} & \underline{25.4} & 25.9 & \underline{5.3} & \textbf{39.1} & 42.6 & \\
    
    \midrule
    \rowcolor{gray!20}
    \multicolumn{11}{c}{\textbf{Qwen2.5-7B}} \\
    \midrule

    Base & -- & 81.6 & 57.4 & 52.2 & 22.1 & 26.4 & 7.3 & 35.5 & 40.4 & - \\
    GRPO & Dense & \underline{91.0} & \textbf{73.8} & \textbf{63.9} & \textbf{37.1} & \textbf{39.1} & \textbf{14.5} & \underline{52.5} & \textbf{53.1} & - \\
    
    \cmidrule(lr){1-11}
    
    GRPO & \multirow{2}{*}{w/ R-KV}
    & 88.0 & 64.2 & 57.1 & 23.2 
    & 30.7 & 8.8 & 43.8 & 45.1 & \multirow{4}{*}{39.4\%} \\
    
    $\hookrightarrow$ + Sparse-RL & 
    & \textbf{91.5} & \underline{72.0} & \underline{61.8} & \underline{32.7} & \underline{36.6} & \underline{12.5} & \textbf{52.7} & \underline{51.4} & \\
    
    \cmidrule(lr){1-10}
    
    GRPO & \multirow{2}{*}{w/ SnapKV}
    & 73.4 & 54.6 & 46.0 & 16.9 & 19.0 & 2.6 & 25.2 & 34.0 & \\
    
    $\hookrightarrow$ + Sparse-RL & 
    & 90.1 & 71.4 & 60.8 & 26.8 & 32.9 & 10.2 & 47.6 & 48.5 & \\

    \bottomrule
  \end{tabular}
  \caption{Main results on 7 mathematical reasoning benchmarks. We compare our \method framework (instantiated with R-KV and SnapKV) against the uncompressed Dense-GRPO baseline and the Naive sparse rollout baseline. The ``Toks. saving'' column indicates the reduction in KV cache storage compared to the dense rollout baseline during training. The top two performances are highlighted in \textbf{bold} and \underline{underlined}.}
  \label{tab:main_result}
\end{table*}

%% file: table/bonus.tex
\begin{table}[t]
  \setlength{\tabcolsep}{4pt}
  \begin{center}
    \scalebox{0.65}{
    \begin{tabular} {lccccc}
      \toprule
      Model & GSM8K & MATH500 & Gaokao & Minerva & Olympiad \\
      \midrule
      \rowcolor{gray!20}
      \multicolumn{6}{c}{Qwen2.5-1.5B} \\
      \midrule
      GRPO (Dense) & 73.4 & 50.8 & 37.7 & 12.5 & 14.2 \\
      $\hookrightarrow$ + \method (R-KV) & \textbf{74.6} & \textbf{54.0} & \textbf{42.9} & \textbf{15.4} & \textbf{18.1}
\\

      \midrule
      \rowcolor{gray!20}
      \multicolumn{6}{c}{Qwen2.5-3B} \\
      \midrule
      GRPO (Dense) & \textbf{85.1} & 54.2 & 44.2 & 16.5 & 17.3 \\
      $\hookrightarrow$ + \method (R-KV) & 84.1 & \textbf{61.8} & \textbf{49.1} & \textbf{22.1} & \textbf{23.3}
\\

      \midrule
      \rowcolor{gray!20}
      \multicolumn{6}{c}{Qwen2.5-7B} \\
      \midrule
      GRPO (Dense) & 91.2 & 71.6 & 55.3 & \textbf{33.8} & 33.0 \\
      $\hookrightarrow$ + \method (R-KV) & \textbf{91.4} & \textbf{71.8} & \textbf{59.0} & 30.9 & \textbf{33.5}
\\

      \bottomrule
    \end{tabular}
    }
    \caption{The performance of models trained by GRPO + \method (with R-KV) and GRPO-Dense in the sparse inference scenario by R-KV compression.}
    \label{tab:bonus}
  \end{center}
\end{table}

%% file: latex/6.conclusion.tex
This paper investigates the critical memory bottleneck issue caused by key-value (KV) cache storage for rollout stage in reinforcement learning. We identify the policy mismatch between sparse sampler and dense learner as the root cause of training crashes under sparse rollout pattern. To address this, we propose \method, a memory-efficient framework that enables sparse RL rollout. We integrate Sparsity-Aware Rejection Sampling and Importance-based Reweighting to ensure stable training. Experiments validate \method's effectiveness, achieving performance on par with dense baselines under strict KV budget constraints while demonstrating robust resilience across varying model scales and compression methods.

%% file: latex/7.limitations.tex
\vpara{Task Generalization.} While \method demonstrates robust performance on reasoning-intensive tasks (e.g., mathematics) with verifiable binary rewards, its applicability to open-ended generation tasks—such as creative writing or general instruction following—remains under-explored. In these scenarios, the definition of anomalous tokens is more ambiguous, and the attention patterns may differ from the redundancy often found in Chain-of-Thought reasoning. Future work is needed to verify whether the distributional correction mechanisms of \method generalize effectively to these broader domains.

\vpara{Sampling Efficiency Trade-off.} \method relies on a strict rejection sampling mechanism to filter out trajectories corrupted by compression-induced information loss. Although this ensures training stability, it inherently introduces a trade-off in sampling efficiency. In cases where the compression budget is extremely aggressive or the policy mismatch is severe, the rejection rate may increase, leading to wasted computational resources on discarded trajectories. Optimizing this trade-off, perhaps through token-level correction during generation rather than sequence-level rejection, remains a direction for future improvement.

%% file: latex/8.appendix.tex
\section{Compression Configuration}
\label{sec:training}

The hyperparameters for R-KV and SnapKV include: (1) $B_{\text{budget}}$, the budget size of cache to store retained KV tokens; (2) $B_{\text{buffer}}$, the size of a buffer for newly generated text tokens. After the model generates each fixed-length text segment in the buffer, the KV cache compression is executed. (3) $\alpha$, the last few tokens that are always retained in the cache as observation tokens. (4) $\lambda$ (only in R-KV), controls the trade-off between prioritizing important tokens and reducing redundant tokens. In our experiments, $B_{\text{budget}}$ = 512, $B_{\text{buffer}}$ = 128, $\alpha$ = 8, $\lambda$ = 0.1, with the latter three parameters following the settings in the R-KV paper~\cite{cai2025r}.

\section{Benchmark Statistics}
\label{sec:bench}

The detailed statistical information of the 7 mathematical benchmarks is shown in Table \ref{tab:bench}.
\input{table/bench}

\section{Dynamics of Rejection Rate and Clip Ratio}
\label{sec:ratio}
We monitor the rejection rate and the policy gradient clipping ratio, using Qwen2.5-3B trained by GRPO + \method (with R-KV).

\begin{figure}[t]
  \centering
  \includegraphics[width=1.0\columnwidth]{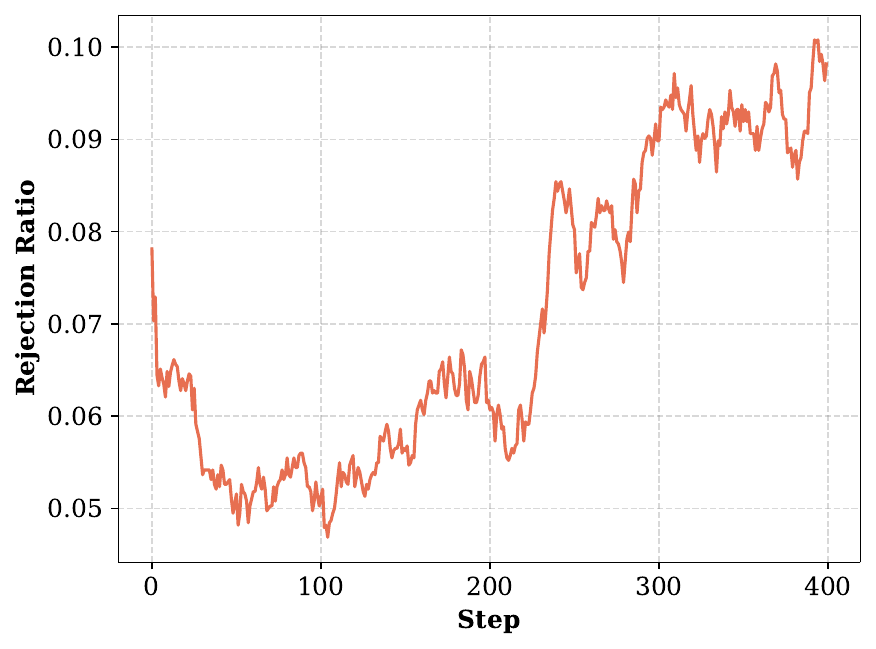}
  \caption{Dynamics of rejection rate on Qwen2.5-3B trained by GRPO + \method (with R-KV).}
 \label{fig:reject_rate}
\end{figure}

\begin{figure}[t]
  \centering
  \includegraphics[width=1.0\columnwidth]{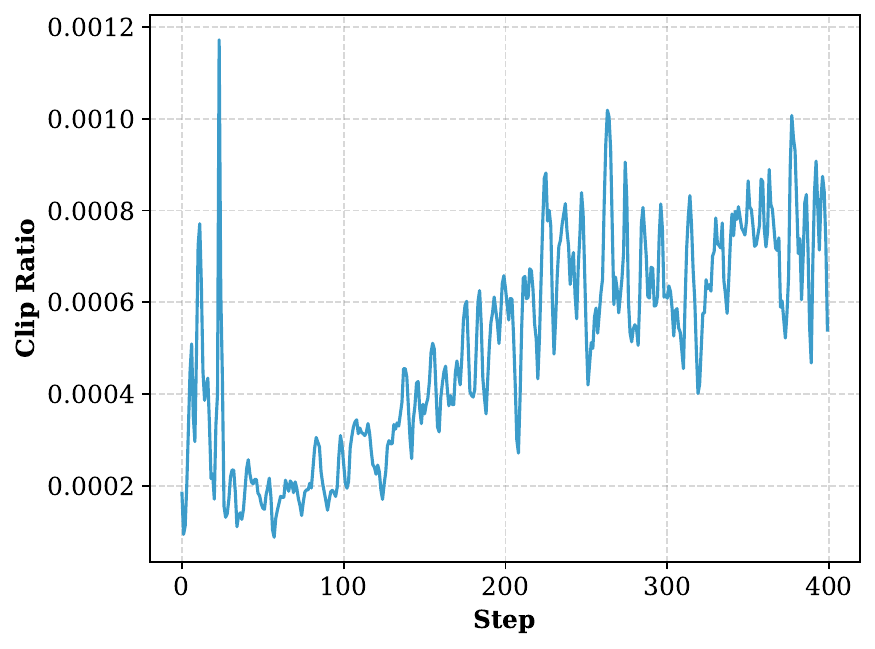}
  \caption{Dynamics of policy gradient clip ratio on Qwen2.5-3B trained by GRPO + \method (with R-KV).}
 \label{fig:clip_ratio}
\end{figure}

Figures \ref{fig:reject_rate} and \ref{fig:clip_ratio} illustrate the detailed training dynamics on Qwen2.5-3B. The rejection ratio (Figure \ref{fig:reject_rate}) fluctuates between roughly 0.05 and 0.11 throughout the training process, with an average of 0.07. This indicates that the vast majority of trajectories generated by the sparse policy satisfy the consistency constraints required by our rejection sampling mechanism. Meanwhile, the clip ratio (Figure \ref{fig:clip_ratio}) remains negligible in the order of $10^{-4}$, confirming that our Importance-Based Reweighting successfully corrects off-policy bias, keeping policy updates well-contained within a stable trust region throughout training.

\section{Preliminaries of GRPO}
\label{sec:preliminaries}

Group Relative Policy Optimization (GRPO), introduced by ~\cite{shao2024deepseekmath}, is a variant of the Proximal Policy Optimization (PPO) algorithm specifically designed for LLMs. The key innovation of GRPO is the elimination of the value model (critic), which significantly reduces computational
overhead. 

GRPO estimates the advantage of a response by comparing its reward against the relative performance within a group of samples generated from the same prompt. For a given input prompt $x$, the model generates $G$ responses $\{o_1, o_2, \dots, o_G\}$. Each response $o_i$ is assigned a reward $r_i$ (through a rule-based verifier or a reward model). The advantage $\hat{A}_i$ for each response $o_i$ is calculated as:
\begin{equation}
\hat{A}_i = \frac{r_i - \text{mean}(r_1, \dots, r_G)}{\text{std}(r_1, \dots, r_G)}
\end{equation}

Based on this , GRPO adopts a clipped surrogate loss similar to PPO:
\begin{equation}
\resizebox{0.99\linewidth}{!}{
$
\begin{aligned}
\mathcal{J}_{\text{GRPO}}(\theta) = \mathbb{E}_{\substack{x \sim \mathcal{D} \\ o_i \sim \pi_{\theta_{\text{old}}}}} \Bigg[ \frac{1}{G} & \sum_{i=1}^G \frac{1}{|o_i|} \sum_{t=1}^{|o_i|} \Big(\min ( w_{i,t}(\theta) \hat{A}_i, \\
& \text{clip}(w_{i,t}(\theta), 1-\epsilon, 1+\epsilon) \hat{A}_i) \Big) \Bigg]
\end{aligned}
$}
\end{equation}
where $w_{i,t}(\theta) = \frac{\pi_\theta(o_{i,t}|x, o_{i,<t})}{\pi_{\theta_{\text{old}}}(o_{i,t}|x,o_{i,<t})}$ is the importance sampling weight (policy ratio), $\epsilon$ is the clipping threshold.

\section{Derivation of \method Objective Function}
\label{sec:derivation}

According to section \ref{subsec:optimization}, the objective is:
\begin{equation}
\label{eq:objective}
\resizebox{0.99\linewidth}{!}{
$
\begin{aligned}
\mathcal{J}_{\text{Sparse-RL}}(\theta) = 
\mathbb{E}_{\substack{x \sim \mathcal{D} \\ o_i \sim \pi_{\theta_{\text{sparse}}}}} 
& \Bigg[ \frac{1}{G} \sum_{i=1}^{G} M^{\text{RS}}(o_i) \cdot \frac{1}{|o_i|} \sum_{t=1}^{|o_i|} \xi_{i,t} \\ 
& \cdot \min \Big( w_{i,t}(\theta) \hat{A}_{i},  \text{clip}\big(w_{i,t}(\theta), 1-\epsilon, 1+\epsilon\big) \hat{A}_{i} \Big)  \Bigg]
\end{aligned}
$}
\end{equation}

The gradient derivation of Eq.\ref{eq:objective} is (clipping is omitted for brevity):
\begin{equation}
\label{eq:gradient_derivation}
\resizebox{1.0\linewidth}{!}{
$
\begin{aligned}
\nabla_\theta \mathcal{J}_{\text{CaRPO}}(\theta) &= \nabla_\theta \mathbb{E}_{\substack{x \sim \mathcal{D} \\ o_i \sim \pi_{\theta_{\text{sparse}}}}} \left[ \frac{1}{G} \sum_{i=1}^{G}  \frac{M^{\text{RS}}(o_i)}{|o_i|} \sum_{t=1}^{|o_i|} \xi_{i,t} \cdot w_{i,t}(\theta) \hat{A}_{i} \right] \\ \\
&= \mathbb{E}_{\substack{x \sim \mathcal{D} \\ o_i \sim \pi_{\theta_{\text{sparse}}}}} \left[ \frac{1}{G} \sum_{i=1}^{G} \frac{M^{\text{RS}}(o_i)}{|o_i|} \sum_{t=1}^{|o_i|} \xi_{i,t} \cdot  \nabla_\theta w_{i,t}(\theta) \hat{A}_{i} \right]
\end{aligned}
$}
\end{equation}

By applying the chain rule to the policy ratio $w_{i,t}(\theta) = \frac{\pi_\theta(o_{i,t}|x, o_{i,<t})}{\pi_{\theta_{\text{old}}}(o_{i,t}|x,o_{i,<t})}$, the gradient term can be expanded:
\begin{equation}
\resizebox{1.0\linewidth}{!}{
$
\nabla_\theta w_{i,t}(\theta) = \frac{\pi_\theta(o_{i,t}|x,o_{i,<t})}{\pi_{\theta_{\text{old}}}(o_{i,t}|x,o_{i,<t})} \nabla_\theta \log \pi_\theta(o_{i,t}|x,o_{i,<t})
$}
\end{equation}

Substituting this back into Eq. \ref{eq:gradient_derivation}, we obtain the final gradient update:
\begin{equation}
\label{eq:final_gradient}
\resizebox{0.95\linewidth}{!}{
$
\begin{aligned}
\nabla_\theta \mathcal{J}_{\text{Sparse-RL}} = \operatorname*{\mathbb{E}}_{\substack{x \sim \mathcal{D} \\ o_i \sim \pi_{\theta_{\text{sparse}}}}} 
& \Bigg[ \frac{1}{G} \sum_{i=1}^{G} M^{\text{RS}}(o_i) \cdot \frac{\hat{A}_{i}}{|o_i|} \\
& \cdot \sum_{t=1}^{|o_i|} \Big( \xi_{i,t} \cdot w_{i,t}(\theta)\nabla_\theta \log \pi_\theta(o_{i,t}|x,o_{i,<t}) \Big) \Bigg]
\end{aligned}
$}
\end{equation}

Because we have:
\begin{equation}
\label{eq:mid}
\resizebox{0.95\linewidth}{!}{
$
\begin{aligned}
\xi_{i,t} \cdot w_{i,t}(\theta) = & \frac{\pi_{\theta_\text{old}}(o_{i,t} | x, o_{i,<t})}{\pi_{\theta_\text{sparse}}(o_{i,t} | x, o_{i,<t})} \cdot \frac{\pi_\theta(o_{i,t}|x, o_{i,<t})}{\pi_{\theta_{\text{old}}}(o_{i,t}|x,o_{i,<t})} \\ \\
&= \frac{\pi_\theta(o_{i,t}|x, o_{i,<t})}{\pi_{\theta_{\text{sparse}}}(o_{i,t}|x,o_{i,<t})}
\end{aligned}
$}
\end{equation}

Substituting Eq. \ref{eq:mid} into Eq. \ref{eq:final_gradient}, the gradient becomes:
\begin{equation}
\resizebox{0.95\linewidth}{!}{
$
\begin{aligned}
\nabla_\theta \mathcal{J}_{\text{Sparse-RL}} = \operatorname*{\mathbb{E}}_{\substack{x \sim \mathcal{D} \\ o_i \sim \pi_{\theta_{\text{sparse}}}}} 
& \Bigg[ \frac{1}{G} \sum_{i=1}^{G} M^{\text{RS}}(o_i) \cdot \frac{\hat{A}_{i}}{|o_i|} \\
& \cdot \sum_{t=1}^{|o_i|} \Big(\frac{\pi_\theta(o_{i,t}|x, o_{i,<t})}{\pi_{\theta_{\text{sparse}}}(o_{i,t}|x,o_{i,<t})}\nabla_\theta \log \pi_\theta(o_{i,t}|x,o_{i,<t}) \Big) \Bigg]
\end{aligned}
$}
\end{equation}

\onecolumn
\section{Example of Anomalous Sequence}
\label{sec:example}

\input{table/samples}

%% file: table/bench.tex
\begin{table}[h]
\centering
\scalebox{0.85}{
\begin{tabular}{lp{5cm}c}
\toprule
\textbf{Benchmark} & \textbf{Description} & \textbf{Size} \\
\midrule
GSM8K & High quality linguistically diverse grade school math word problems. & 1319 \\
MATH500 & A subset from the MATH benchmark that OpenAI created in their \textit{Let's Verify Step by Step} paper. & 500 \\
Minerva & A curated subset of Minerva targeting advanced STEM reasoning from undergraduate math and science exams. & 272 \\
Gaokao & Math questions from Chinese college entrance examinations. & 385 \\
Olympiad & Olympiad-level mathematics and physics competitions. & 675 \\
AIME24 & Problems from the 2024 American Invitational Mathematics Examination. & 30 \\
AMC23 & Problems from the 2023 American Mathematics Competition. & 40 \\
\bottomrule
\end{tabular}
}
\caption{Description of benchmark}
\label{tab:bench}
\end{table}

%% file: table/samples.tex
\noindent
\begin{minipage}[t]{0.48\columnwidth}
\begin{tcolorbox}[
  title={Prompt \& Dense Rollout (Correct)},
  colback=gray!5,
  colframe=green!50!black
]
\small
\textbf{Prompt:} 

The expression $2\cdot 3 \cdot 4\cdot 5+1$ is equal to 121, since multiplication is carried out before addition. However, we can obtain values other than 121 for this expression if we are allowed to change it by inserting parentheses. For example,
\[
(2\cdot (3\cdot 4)) \cdot (5+1) = 144.
\]
In total, how many values can be obtained? \\[0.5em]

\textbf{Dense Rollout (Correct):}

We can reason step by step as follows:

1. We have the expression $2 \cdot 3 \cdot 4 \cdot 5 + 1$.

2. We can insert parentheses to change the order of operations and obtain different values.

3. Let's consider each possible position for inserting parentheses: \\
(a) Between 2 and 3: $(2 \cdot 3) \cdot 4 \cdot 5 + 1$ \\
(b) Between 3 and 4: $2 \cdot (3 \cdot 4) \cdot 5 + 1$ \\
(c) Between 4 and 5: $2 \cdot 3 \cdot (4 \cdot 5) + 1$ \\
(d) Between 5 and 1: $2 \cdot 3 \cdot 4 \cdot (5 + 1)$ \\

4. We can calculate the value of each modified expression: \\
(a) $(2 \cdot 3) \cdot 4 \cdot 5 + 1 = 6 \cdot 4 \cdot 5 + 1 = 120 + 1 = 121$ \\
(b) $2 \cdot (3 \cdot 4) \cdot 5 + 1 = 2 \cdot 12 \cdot 5 + 1 = 120 + 1 = 121$ \\
(c) $2 \cdot 3 \cdot (4 \cdot 5) + 1 = 2 \cdot 3 \cdot 20 + 1 = 120 + 1 = 121$ \\
(d) $2 \cdot 3 \cdot 4 \cdot (5 + 1) = 2 \cdot 3 \cdot 4 \cdot 6 = 144$ \\

5. We have found four different values that can be obtained by inserting parentheses in the expression: 121, 121, 121, and 144.

6. Therefore, the total number of values that can be obtained from the expression $2 \cdot 3 \cdot 4 \cdot 5 + 1$ by inserting parentheses is 4. \\
\makebox[\linewidth][r]{\textcolor{green!90!white}{\scalebox{1.4}{\checkmark}}}

\end{tcolorbox}
\end{minipage}
\hfill
\begin{minipage}[t]{0.48\columnwidth}
\begin{tcolorbox}[
  title={Sparse Rollout (Anomalous)},
  colback=gray!5,
  colframe=red!60!black
]
\small
To determine how many distinct values can be obtained from the expression \(2 \cdot 3 \cdot 4 \cdot 5 + 1\) by inserting parentheses, we need to consider all possible ways to distribute multiplication operations across the sequence of numbers and the addition. Let's break this down step by step.

Step 1: List Out All Possible Parenthesizations

We start by listing out all potential combinations of inserting parentheses into the expression \(2 \cdot 3 \cdot 4 \cdot 5 + 1\). We will consider different groupings of the multiplications, but not changing the overall structure of the additions or subtractions, as only the grouping of the multiplications is allowed.

Here are the possible ways:
\begin{enumerate}[nosep, leftmargin=*]
    \item \((2 \cdot 3) \cdot (4 \cdot 5) + 1\)
    \item \(2 \cdot (3 \cdot (4 \cdot 5)) + 1\)
    \item \(2 \cdot (3 \cdot (4 \cdot (5 + 1)))\)
    \item \(((2 \cdot 3) \cdot 4) \cdot 5 + 1\)
    \item \(2 \cdot ((3 \cdot 4) \cdot 5) + 1\)
    \item \(2 \cdot (3 \cdot (4 \cdot (5 + 1)))\)
    \item \(2 \cdot 3 \cdot 4 \cdot (5 + 1) + 1\)
    \item \((2 \cdot 3 \cdot (4 \cdot 5)) + 1\)
    \item \(2 \cdot ((3 \cdot 4) \cdot (5 + 1)) + 1\)
\end{enumerate}

Step 2: Calculate Each Parenthesized Expression

Now, let's calculate each of these expressions individually to find out their respective values:
\begin{enumerate}[nosep, leftmargin=*]
    \item \((2 \cdot 3) \cdot (4 \cdot 5) + 1 = 6 \cdot 20 + 1 = 120 + 1 = 121\)
    \item \(2 \cdot (3 \cdot (4 \cdot 5)) + 1 = 2 \cdot (3 \cdot 20) + 1 = 2 \cdot 60 + 1 = 120 + 1 = 121\)
    \item \(2 \cdot (3 \cdot (4 \cdot (5 + 1))) = 2 \cdot (3 \cdot (4 \cdot 6)) + 1 = 2 \cdot (3 \cdot 24) + 1 = 2 \cdot 72 + 1 = 144 + 1 = 145\)
    \item \(((2 \cdot 3) \cdot 4) \cdot 5 + 1 = (6 \cdot 4) \cdot 5 + 1 = 24 \cdot 5 + 1 = 120 + 1 = 121\)
    \item \(2 \cdot ((3 \cdot 4) \cdot 5) + 1 = 2 \cdot (12 \cdot 5) + 1 = 2 \cdot 60 + 1 = 120 + 1 = 121\)
    \item \(2 \cdot (3 \cdot (4 \cdot (5 + 1))) = 2 \cdot (3 \cdot 24) + 1 = 2 \cdot 72 + 1 = \textcolor{red}{141 + 1 = 141 + 1 = 141 + 1 = 141 + 1 \dots}  \) 
\end{enumerate}

\makebox[\linewidth][r]{\textcolor{red}{\textit{[Infinite repetition]}}}

\end{tcolorbox}
\end{minipage}

\captionof{figure}{The information loss caused by KV cache compression may generate anomalous sequences (e.g., infinite repetition) during rollout phase. We provide a concrete example sampled during the training of Qwen2.5-3B with R-KV compression (budget=512).}

%% file: custom.bib
@article{jaech2024openai,
  title={Openai o1 system card},
  author={Jaech, Aaron and Kalai, Adam and Lerer, Adam and Richardson, Adam and El-Kishky, Ahmed and Low, Aiden and Helyar, Alec and Madry, Aleksander and Beutel, Alex and Carney, Alex and others},
  journal={arXiv preprint arXiv:2412.16720},
  year={2024}
}

@article{guo2025deepseek,
  title={Deepseek-r1: Incentivizing reasoning capability in llms via reinforcement learning},
  author={Guo, Daya and Yang, Dejian and Zhang, Haowei and Song, Junxiao and Zhang, Ruoyu and Xu, Runxin and Zhu, Qihao and Ma, Shirong and Wang, Peiyi and Bi, Xiao and others},
  journal={arXiv preprint arXiv:2501.12948},
  year={2025}
}

@article{chen2025towards,
  title={Towards reasoning era: A survey of long chain-of-thought for reasoning large language models},
  author={Chen, Qiguang and Qin, Libo and Liu, Jinhao and Peng, Dengyun and Guan, Jiannan and Wang, Peng and Hu, Mengkang and Zhou, Yuhang and Gao, Te and Che, Wanxiang},
  journal={arXiv preprint arXiv:2503.09567},
  year={2025}
}

@article{ouyang2022training,
  title={Training language models to follow instructions with human feedback},
  author={Ouyang, Long and Wu, Jeffrey and Jiang, Xu and Almeida, Diogo and Wainwright, Carroll and Mishkin, Pamela and Zhang, Chong and Agarwal, Sandhini and Slama, Katarina and Ray, Alex and others},
  journal={Advances in neural information processing systems},
  volume={35},
  pages={27730--27744},
  year={2022}
}

@article{shao2024deepseekmath,
  title={Deepseekmath: Pushing the limits of mathematical reasoning in open language models},
  author={Shao, Zhihong and Wang, Peiyi and Zhu, Qihao and Xu, Runxin and Song, Junxiao and Bi, Xiao and Zhang, Haowei and Zhang, Mingchuan and Li, YK and Wu, Yang and others},
  journal={arXiv preprint arXiv:2402.03300},
  year={2024}
}

@misc{team2025qwq,
  title={Qwq-32b: Embracing the power of reinforcement learning},
  author={Team, Qwen},
  year={2025},
  publisher={March}
}

@article{liu2025code,
  title={Code-r1: Reproducing r1 for code with reliable rewards},
  author={Liu, Jiawei and Zhang, Lingming},
  journal={arXiv preprint arXiv:2503.18470},
  volume={3},
  year={2025}
}

@article{shao2025beat,
  title={Beat the long tail: Distribution-Aware Speculative Decoding for RL Training},
  author={Shao, Zelei and Srivatsa, Vikranth and Srivastava, Sanjana and Wu, Qingyang and Ariyak, Alpay and Wu, Xiaoxia and Patel, Ameen and Wang, Jue and Liang, Percy and Dao, Tri and others},
  journal={arXiv preprint arXiv:2511.13841},
  year={2025}
}

@article{wang2025rlhfspec,
  title={RLHFSpec: Breaking the Efficiency Bottleneck in RLHF Training via Adaptive Drafting},
  author={Wang, Siqi and Yang, Hailong and Zhu, Junjie and Wang, Xuezhu and Xu, Yufan and Qian, Depei},
  journal={arXiv preprint arXiv:2512.04752},
  year={2025}
}

@article{zhao2025accelerating,
  title={Accelerating Large-Scale Reasoning Model Inference with Sparse Self-Speculative Decoding},
  author={Zhao, Yilong and Tang, Jiaming and Zhu, Kan and Ye, Zihao and Chang, Chi-Chih and Lin, Chaofan and Park, Jongseok and Xiao, Guangxuan and Abdelfattah, Mohamed S and Gao, Mingyu and others},
  journal={arXiv preprint arXiv:2512.01278},
  year={2025}
}

@article{zhang2023h2o,
  title={H2o: Heavy-hitter oracle for efficient generative inference of large language models},
  author={Zhang, Zhenyu and Sheng, Ying and Zhou, Tianyi and Chen, Tianlong and Zheng, Lianmin and Cai, Ruisi and Song, Zhao and Tian, Yuandong and R{\'e}, Christopher and Barrett, Clark and others},
  journal={Advances in Neural Information Processing Systems},
  volume={36},
  pages={34661--34710},
  year={2023}
}

@article{xiao2023efficient,
  title={Efficient streaming language models with attention sinks},
  author={Xiao, Guangxuan and Tian, Yuandong and Chen, Beidi and Han, Song and Lewis, Mike},
  journal={arXiv preprint arXiv:2309.17453},
  year={2023}
}

@article{li2024snapkv,
  title={Snapkv: Llm knows what you are looking for before generation},
  author={Li, Yuhong and Huang, Yingbing and Yang, Bowen and Venkitesh, Bharat and Locatelli, Acyr and Ye, Hanchen and Cai, Tianle and Lewis, Patrick and Chen, Deming},
  journal={Advances in Neural Information Processing Systems},
  volume={37},
  pages={22947--22970},
  year={2024}
}

@article{cai2024pyramidkv,
  title={Pyramidkv: Dynamic kv cache compression based on pyramidal information funneling},
  author={Cai, Zefan and Zhang, Yichi and Gao, Bofei and Liu, Yuliang and Li, Yucheng and Liu, Tianyu and Lu, Keming and Xiong, Wayne and Dong, Yue and Hu, Junjie and others},
  journal={arXiv preprint arXiv:2406.02069},
  year={2024}
}

@article{feng2024ada,
  title={Ada-kv: Optimizing kv cache eviction by adaptive budget allocation for efficient llm inference},
  author={Feng, Yuan and Lv, Junlin and Cao, Yukun and Xie, Xike and Zhou, S Kevin},
  journal={arXiv preprint arXiv:2407.11550},
  year={2024}
}

@article{cai2025r,
  title={R-KV: Redundancy-aware KV Cache Compression for Training-Free Reasoning Models Acceleration},
  author={Cai, Zefan and Xiao, Wen and Sun, Hanshi and Luo, Cheng and Zhang, Yikai and Wan, Ke and Li, Yucheng and Zhou, Yeyang and Chang, Li-Wen and Gu, Jiuxiang and others},
  journal={arXiv preprint arXiv:2505.24133},
  year={2025}
}

@article{ge2023model,
  title={Model tells you what to discard: Adaptive kv cache compression for llms},
  author={Ge, Suyu and Zhang, Yunan and Liu, Liyuan and Zhang, Minjia and Han, Jiawei and Gao, Jianfeng},
  journal={arXiv preprint arXiv:2310.01801},
  year={2023}
}

@article{liu2023scissorhands,
  title={Scissorhands: Exploiting the persistence of importance hypothesis for llm kv cache compression at test time},
  author={Liu, Zichang and Desai, Aditya and Liao, Fangshuo and Wang, Weitao and Xie, Victor and Xu, Zhaozhuo and Kyrillidis, Anastasios and Shrivastava, Anshumali},
  journal={Advances in Neural Information Processing Systems},
  volume={36},
  pages={52342--52364},
  year={2023}
}

@article{hooper2024kvquant,
  title={Kvquant: Towards 10 million context length llm inference with kv cache quantization},
  author={Hooper, Coleman and Kim, Sehoon and Mohammadzadeh, Hiva and Mahoney, Michael W and Shao, Yakun S and Keutzer, Kurt and Gholami, Amir},
  journal={Advances in Neural Information Processing Systems},
  volume={37},
  pages={1270--1303},
  year={2024}
}

@inproceedings{zhang2024cam,
  title={Cam: Cache merging for memory-efficient llms inference},
  author={Zhang, Yuxin and Du, Yuxuan and Luo, Gen and Zhong, Yunshan and Zhang, Zhenyu and Liu, Shiwei and Ji, Rongrong},
  booktitle={Forty-first international conference on machine learning},
  year={2024}
}

@article{sun2024shadowkv,
  title={Shadowkv: Kv cache in shadows for high-throughput long-context llm inference},
  author={Sun, Hanshi and Chang, Li-Wen and Bao, Wenlei and Zheng, Size and Zheng, Ningxin and Liu, Xin and Dong, Harry and Chi, Yuejie and Chen, Beidi},
  journal={arXiv preprint arXiv:2410.21465},
  year={2024}
}

@article{zeng2025simplerl,
  title={Simplerl-zoo: Investigating and taming zero reinforcement learning for open base models in the wild},
  author={Zeng, Weihao and Huang, Yuzhen and Liu, Qian and Liu, Wei and He, Keqing and Ma, Zejun and He, Junxian},
  journal={arXiv preprint arXiv:2503.18892},
  year={2025}
}

@article{hendrycks2021measuring,
  title={Measuring mathematical problem solving with the math dataset},
  author={Hendrycks, Dan and Burns, Collin and Kadavath, Saurav and Arora, Akul and Basart, Steven and Tang, Eric and Song, Dawn and Steinhardt, Jacob},
  journal={arXiv preprint arXiv:2103.03874},
  year={2021}
}

@article{cobbe2021training,
  title={Training verifiers to solve math word problems},
  author={Cobbe, Karl and Kosaraju, Vineet and Bavarian, Mohammad and Chen, Mark and Jun, Heewoo and Kaiser, Lukasz and Plappert, Matthias and Tworek, Jerry and Hilton, Jacob and Nakano, Reiichiro and others},
  journal={arXiv preprint arXiv:2110.14168},
  year={2021}
}

@article{lewkowycz2022solving,
  title={Solving quantitative reasoning problems with language models},
  author={Lewkowycz, Aitor and Andreassen, Anders and Dohan, David and Dyer, Ethan and Michalewski, Henryk and Ramasesh, Vinay and Slone, Ambrose and Anil, Cem and Schlag, Imanol and Gutman-Solo, Theo and others},
  journal={Advances in neural information processing systems},
  volume={35},
  pages={3843--3857},
  year={2022}
}

@inproceedings{he2024olympiadbench,
  title={Olympiadbench: A challenging benchmark for promoting agi with olympiad-level bilingual multimodal scientific problems},
  author={He, Chaoqun and Luo, Renjie and Bai, Yuzhuo and Hu, Shengding and Thai, Zhen and Shen, Junhao and Hu, Jinyi and Han, Xu and Huang, Yujie and Zhang, Yuxiang and others},
  booktitle={Proceedings of the 62nd Annual Meeting of the Association for Computational Linguistics (Volume 1: Long Papers)},
  pages={3828--3850},
  year={2024}
}

@article{schulman2017proximal,
  title={Proximal policy optimization algorithms},
  author={Schulman, John and Wolski, Filip and Dhariwal, Prafulla and Radford, Alec and Klimov, Oleg},
  journal={arXiv preprint arXiv:1707.06347},
  year={2017}
}

@article{rafailov2023direct,
  title={Direct preference optimization: Your language model is secretly a reward model},
  author={Rafailov, Rafael and Sharma, Archit and Mitchell, Eric and Manning, Christopher D and Ermon, Stefano and Finn, Chelsea},
  journal={Advances in neural information processing systems},
  volume={36},
  pages={53728--53741},
  year={2023}
}

@misc{yao2025offpolicy,
  title = {Your Efficient RL Framework Secretly Brings You Off-Policy RL Training},
  url = {https://fengyao.notion.site/off-policy-rl},
  author = {Yao, Feng and Liu, Liyuan and Zhang, Dinghuai and Dong, Chengyu and Shang, Jingbo and Gao, Jianfeng},
  journal = {Feng Yao's Notion},
  year = {2025},
  month = aug,
}

@misc{liu-li-2025,
  title = {When Speed Kills Stability: Demystifying RL Collapse from the Inference-Training Mismatch},
  url = {https://yingru.notion.site/When-Speed-Kills-Stability-Demystifying-RL-Collapse-from-the-Inference-Training-Mismatch-271211a558b7808d8b12d403fd15edda},
  author = {Jiacai Liu and Yingru Li and Yuqian Fu and Jiawei Wang and Qian Liu and Yu Shen},
  year = {2025},
}

@misc{qwen2025qwen25technicalreport,
      title={Qwen2.5 Technical Report}, 
      author={Qwen and : and An Yang and Baosong Yang and Beichen Zhang and Binyuan Hui and Bo Zheng and Bowen Yu and Chengyuan Li and Dayiheng Liu and Fei Huang and Haoran Wei and Huan Lin and Jian Yang and Jianhong Tu and Jianwei Zhang and Jianxin Yang and Jiaxi Yang and Jingren Zhou and Junyang Lin and Kai Dang and Keming Lu and Keqin Bao and Kexin Yang and Le Yu and Mei Li and Mingfeng Xue and Pei Zhang and Qin Zhu and Rui Men and Runji Lin and Tianhao Li and Tianyi Tang and Tingyu Xia and Xingzhang Ren and Xuancheng Ren and Yang Fan and Yang Su and Yichang Zhang and Yu Wan and Yuqiong Liu and Zeyu Cui and Zhenru Zhang and Zihan Qiu},
      year={2025},
      eprint={2412.15115},
      archivePrefix={arXiv},
      primaryClass={cs.CL}, 
}

@article{dubey2024llama,
  title={The llama 3 herd of models},
  author={Dubey, Abhimanyu and Jauhri, Abhinav and Pandey, Abhinav and Kadian, Abhishek and Al-Dahle, Ahmad and Letman, Aiesha and Mathur, Akhil and Schelten, Alan and Yang, Amy and Fan, Angela and others},
  journal={arXiv e-prints},
  pages={arXiv--2407},
  year={2024}
}

@misc{slime_github,
  author       = {Zilin Zhu and Chengxing Xie and Xin Lv and slime Contributors},
  title        = {slime: An LLM post-training framework for RL Scaling},
  year         = {2025},
  howpublished = {\url{https://github.com/THUDM/slime}},
  note         = {GitHub repository. Corresponding author: Xin Lv},
  urldate      = {2025-06-19}
}

@article{zhang2023evaluating,
  title={Evaluating the performance of large language models on gaokao benchmark},
  author={Zhang, Xiaotian and Li, Chunyang and Zong, Yi and Ying, Zhengyu and He, Liang and Qiu, Xipeng},
  journal={arXiv preprint arXiv:2305.12474},
  year={2023}
}

@article{zhang2025agentrl,
  title={AgentRL: Scaling Agentic Reinforcement Learning with a Multi-Turn, Multi-Task Framework},
  author={Zhang, Hanchen and Liu, Xiao and Lv, Bowen and Sun, Xueqiao and Jing, Bohao and Iong, Iat Long and Hou, Zhenyu and Qi, Zehan and Lai, Hanyu and Xu, Yifan and others},
  journal={arXiv preprint arXiv:2510.04206},
  year={2025}
}

@misc{fu2025areal,
      title={AReaL: A Large-Scale Asynchronous Reinforcement Learning System for Language Reasoning},
      author={Wei Fu and Jiaxuan Gao and Xujie Shen and Chen Zhu and Zhiyu Mei and Chuyi He and Shusheng Xu and Guo Wei and Jun Mei and Jiashu Wang and Tongkai Yang and Binhang Yuan and Yi Wu},
      year={2025},
      eprint={2505.24298},
      archivePrefix={arXiv},
      primaryClass={cs.LG},
      url={https://arxiv.org/abs/2505.24298},
}

@article{zheng2025group,
  title={Group sequence policy optimization},
  author={Zheng, Chujie and Liu, Shixuan and Li, Mingze and Chen, Xiong-Hui and Yu, Bowen and Gao, Chang and Dang, Kai and Liu, Yuqiong and Men, Rui and Yang, An and others},
  journal={arXiv preprint arXiv:2507.18071},
  year={2025}
}

@misc{qin2025seeronlinecontextlearning,
      title={Seer: Online Context Learning for Fast Synchronous LLM Reinforcement Learning}, 
      author={Ruoyu Qin and Weiran He and Weixiao Huang and Yangkun Zhang and Yikai Zhao and Bo Pang and Xinran Xu and Yingdi Shan and Yongwei Wu and Mingxing Zhang},
      year={2025}
}

@article{liu2024kivi,
  title={Kivi: A tuning-free asymmetric 2bit quantization for kv cache},
  author={Liu, Zirui and Yuan, Jiayi and Jin, Hongye and Zhong, Shaochen and Xu, Zhaozhuo and Braverman, Vladimir and Chen, Beidi and Hu, Xia},
  journal={arXiv preprint arXiv:2402.02750},
  year={2024}
}
